\documentclass[10pt,twocolumn]{article}

\usepackage[letterpaper,top=0.66in,bottom=0.72in,left=0.66in,right=0.66in,
            columnsep=0.25in,headsep=0.18in]{geometry}
\usepackage{newtxtext,newtxmath}

\usepackage{amsmath,amssymb}
\usepackage{graphicx}
\usepackage{booktabs,tabularx}
\usepackage{algorithm}
\usepackage{algorithmic}
\usepackage{microtype}
\usepackage{authblk}
\usepackage{caption}
\usepackage{natbib}
\usepackage[hidelinks]{hyperref}

\hypersetup{
  pdftitle={Blind to the Pivotal Vote: Aggregate Independence Metrics Miss Where Verification Actually Helps},
  pdfauthor={Yang Shu}
}
\title{Blind to the Pivotal Vote: Aggregate Independence Metrics Miss Where Verification Actually Helps}
\author[1]{Yang Shu\thanks{Corresponding author: \href{mailto:shuyang@zju.edu.cn}{shuyang@zju.edu.cn}.}}
\affil[1]{Zhejiang University, Hangzhou, China}
\date{\small Preprint --- August 2026}

\begin{document}

\maketitle

\begin{abstract}
LLM judge panels are a standard evaluation tool, but prior work reports highly correlated panel errors: nine judges provide roughly the effective information of two independent ones, and aggregation closes only a small fraction of the gap. A natural remedy -- a signal from a different evidence source, e.g.\ executing a test suite -- produced no distinguishable change in the panel's effective-vote count at scale ($-0.04$, 95\% CI $[-0.10,+0.02]$). Aggregate dependence and conditional decision utility are different questions. Elementary majority arithmetic fixes the affected set for single-ballot substitution: only decisions with a one-vote margin can change. The empirical question is whether panel error rates rise and useful substitutions concentrate there. They do: the entire accuracy gain concentrates on these \emph{pivotal} queries, where it is large (+10.4 to +23.3pp across three headline configurations), and is exactly zero elsewhere. We confirm the pattern across three code benchmarks and four panel sizes (a 9-judge extension and 56 dependent subsampling checks, gain +6.5 to +16.1pp). On HumanEval+/MBPP+, a majority-side replacement rule raises overall accuracy from 82.44\% to 85.62\% while invoking the signal on 16.2\% of queries; signal-only remains stronger at 87.60\%. Thus population-level dependence diagnostics and margin-stratified utility are complementary, and the affected-set characterization yields a call-reduction rule for any specified single-ballot substitution policy.
\end{abstract}

\section{Introduction}

Multi-model LLM judge panels are increasingly used to evaluate model outputs at scale, on the intuition that pooling several independent judgments should be more reliable than trusting any single one. \citet{kohli2026nine} recently complicated this intuition: across nine frontier judges from seven model families, panel errors are so correlated that the panel carries roughly the statistical information of two truly independent judges, and even an oracle-calibrated aggregation rule closes at most about eleven percent of the resulting reliability gap. Their result suggests that simply adding more LLM judges can offer diminishing returns when the models share blind spots.

A natural response is to add a vote derived from another procedure. In code generation, a partial unit-test execution can supply an imperfect pass/fail ballot. Following \citeauthor{kohli2026nine}'s effective-votes methodology, seven LLM judges blindly score candidate solutions and the verification signal replaces one judge. A pilot suggested an effective-vote increase, but its 98\%-accurate signal had only five errors from one solver and could not support a general conclusion. At ten times the scale, a less accurate signal produced no distinguishable change in effective votes (95\% CI crossing zero), although panel accuracy still improved.

The improvement becomes interpretable after stratifying queries by vote margin. When at least five of seven judges agree, single-ballot substitution changes zero decisions and gains 0.0000 accuracy. On 4--3 votes, the same signal improves accuracy by more than ten percentage points. The boundary follows from majority arithmetic: replacing one voter can change the vote total by at most one, so it can flip the outcome only when the original tally is one vote from a tie. We call such a query \emph{pivotal}. The empirical claim is not that most panel errors occur there -- only 199 of 569 errors do in the main setting -- but that panel error \emph{rates} rise as the margin narrows and all decision changes and accuracy gains from single-ballot substitution are confined there.

\begin{figure*}[!t]
\centering
\includegraphics[width=\textwidth]{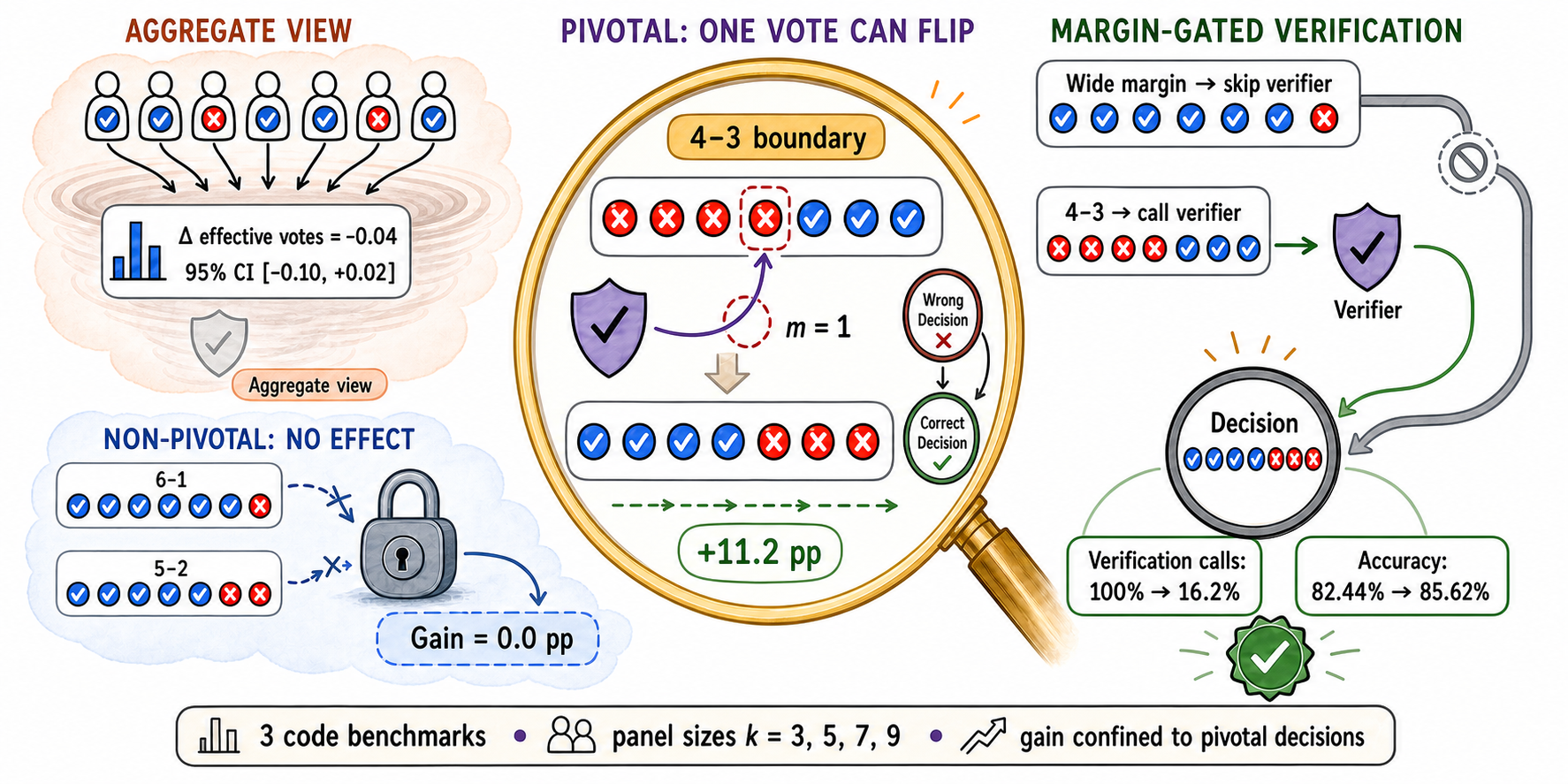}
\caption{Overview of the aggregate blind spot and pivotal-vote mechanism. On HumanEval+/MBPP+ with $k{=}7$, random replacement gains $+11.2$pp within the pivotal region; majority-side gating invokes the verifier on 16.2\% of queries and improves overall accuracy from 82.44\% to 85.62\%. Icons illustrate the mechanism; experiments span three code benchmarks and $k\in\{3,5,7,9\}$.}
\label{fig:overview}
\end{figure*}

This has two consequences. First, an aggregate statistic like the effective-vote count averages a population that is 84\% unaffected by substitution and 16\% strongly affected; the average can therefore obscure a reliable conditional effect. Second, once a substitution rule has been specified, verification can be skipped outside the pivotal region without changing that rule's predictions. This reduces verifier-call frequency to 12--27\% in our experiments. It does not establish a total-system cost advantage, and universal execution is already inexpensive for the test-based signal studied here. Figure~\ref{fig:overview} summarizes this aggregate-to-pivotal-to-gating logic.

Our contributions are: (1) we separate a structural statement from an empirical one: elementary majority-vote arithmetic identifies the exact affected set for single-ballot substitution (Propositions~1--2), while measurements across three code benchmarks and four panel sizes show that error rates rise and substitution gains concentrate there; (2) using the same $n_{\mathrm{eff}}$ statistic as \citet{kohli2026nine}, we show that population-level dependence and margin-conditional substitution utility are complementary diagnostics on the same data (Table~\ref{tab:neff}); and (3) we evaluate margin gating with uniformly random, fixed, and majority-side replacement rules, showing that gating preserves each rule's universal-substitution predictions while the replacement rule materially affects accuracy (Table~\ref{tab:baselines}). A secondary, exploratory analysis associates part of the pivotal-region error pattern with panel composition (Section~5).

\section{Related Work}

\paragraph{Correlated errors in LLM judge panels.} \citet{kohli2026nine} is the closest prior work: it introduces the effective-votes framework we build on, $n_{\mathrm{eff}} = k / (1 + (k-1)\bar\phi)$, a Kish-style effective sample size over the mean pairwise correlation $\bar\phi$ of judges' (or judge-plus-signal's) binary error vectors for panel size $k$; we use it exactly as defined there. Their diagnosis -- correlated judges, not weak aggregation, is the bottleneck -- motivates our search for a signal drawn from a different source of evidence, but their framework, as we show, is not designed to detect value concentrated in a small, identifiable query subset. \citet{kuai2026independent} audit behavioral entanglement among LLMs and reweight verifier ensembles, but only among LLMs, not non-LLM signals or $n_{\mathrm{eff}}$ itself. \citet{kwok2026llmverifier} optimize a single verifier's scoring granularity, a complementary but orthogonal problem to multi-judge panels. \citet{verga2024replacing} propose PoLL, a panel of diverse smaller LLM judges cheaper and less self-biased than one strong judge, motivating multi-judge panels as we assume here, but without conditioning aggregation on vote margin.

\paragraph{Margin-gated verification as engineering practice.} \citet{akinfaderin2026fregelogic}'s SemEval system already invokes a symbolic (Z3) verifier when a five-model ensemble splits 3--2. Thus narrow-margin verification is not our algorithmic invention. We instead isolate the aggregation-rule boundary, test the full margin sweep, distinguish aggregate dependence from conditional utility, and compare replacement rules on substantially more panel decisions. These contributions position our work as a systematic empirical and methodological analysis of the heuristic rather than its first use.

\paragraph{Selective computation and disagreement-based routing.} Our margin gate instantiates the older idea of spending a costly resource where cheap signals indicate it is needed. \citet{seung1992query} select informative points by committee disagreement; selective prediction (\citealp{geifman2017selective}) abstains on low-confidence inputs; learning-to-defer (\citealp{mozannar2020consistent}) routes decisions to an expert; and FrugalGPT (\citealp{chen2023frugalgpt}) uses cost-aware model cascades. Our narrower contribution is to derive the exact affected set for single-ballot substitution under equal-weight majority aggregation and to measure the resulting conditional utility.

\paragraph{Theoretical framing.} Work on multi-judge and multi-agent voting for LLMs is overwhelmingly framed in terms of the Condorcet Jury Theorem, which concerns whether independent, better-than-chance voters make majority voting reliable in aggregate (\citealp{condorcet1785essai}; used explicitly by \citealp{kohli2026nine} and by several majority-voting-harm analyses in adjacent work). We instead draw on the notion of a \emph{pivotal} or decisive voter, formalized in social choice theory by the Banzhaf power index (\citealp{banzhaf1965weighted}), which measures how often a single voter's ballot determines the outcome of a vote. To our knowledge this framing has not previously been applied to LLM judge or ensemble panels; it is a natural fit for our question because it concerns \emph{which} votes can be changed by a single ballot, rather than whether voters are independent on average.

\paragraph{Verifiable-domain benchmarks.} We evaluate on HumanEval+ \citep{liu2023your}, an extended-test-case version of HumanEval \citep{chen2021evaluating}; MBPP+ \citep{liu2023your}, similarly derived from MBPP \citep{austin2021program}; and LiveCodeBench \citep{jain2024livecodebench}, a contamination-resistant benchmark of recent competitive-programming problems that we use as a harder setting than HumanEval+/MBPP+.

\section{The Pivotal-Vote Mechanism}

\paragraph{Setup.} A panel of $k$ judges (here $k \in \{3,5,7,9\}$) blindly scores a candidate solution as correct or incorrect; the panel's decision is the majority vote. For query $i$, let $s_i$ be the number of judges voting ``correct.'' We measure the panel's disagreement with the \emph{margin} $m_i = |2s_i - k|$: $m_i = k$ means unanimity, and for odd $k$ the smallest possible value $m_i = 1$ means the vote is decided by a single ballot (e.g.\ $\lceil k/2 \rceil$ vs.\ $\lfloor k/2 \rfloor$). We call a query with $m_i = 1$ \emph{pivotal}.

\paragraph{Why value is structurally confined to pivotal queries.}

\begin{quote}
\textbf{Proposition 1.} \emph{Let a $k$-judge panel ($k$ odd) predict the label for query $i$ by unweighted majority vote, with $s_i$ of the $k$ votes cast ``correct.'' There exists a choice of which single judge's vote to replace with a verification signal's vote (correct or incorrect) that changes the panel's predicted label if and only if $m_i = |2s_i - k| = 1$; when $m_i \geq 3$, no choice of replaced judge or replacement value can change the predicted label.}
\end{quote}

\emph{Proof.} Let $s_i'$ be the vote total after replacement. Since only one of the $k$ ballots changed, $s_i' \in \{s_i - 1, s_i, s_i+1\}$. The panel predicts the label CORRECT iff the tally exceeds $k/2$. Because $k$ is odd, the decision boundary lies strictly between the two integers $\lfloor k/2 \rfloor$ and $\lceil k/2 \rceil = \lfloor k/2 \rfloor + 1$. A change of at most one vote can move $s_i$ across this boundary only if $s_i \in \{\lfloor k/2\rfloor, \lceil k/2 \rceil\}$, i.e.\ $m_i = |2s_i-k| = 1$, and in that case replacing a vote on the majority side with the opposite value does cross it. For any $s_i$ with $m_i \geq 3$, $s_i$ is at least two votes from the boundary and $s_i'$ cannot cross it regardless of which judge is replaced or what value replaces it. $\blacksquare$ Note this bounds the label the panel \emph{predicts}; whether that prediction is \emph{accurate} against ground truth is a separate, empirical question we test directly below.

This is a property of majority voting under single-ballot substitution, not a claim requiring statistical evidence: it holds by construction for \emph{any} panel and \emph{any} replacement signal, regardless of how accurate that signal is. It immediately implies that the \emph{expected accuracy gain from substitution is exactly zero on every non-pivotal query}, since the panel's output -- and hence its correctness -- cannot change there.

\paragraph{Even panels: an asymmetric extension.} All our experiments use odd $k$, where the vote total alone decides the outcome. Even $k$ requires a tie-breaking convention; we consider the natural one where ties default to ``incorrect'' (decision $=\mathrm{correct}$ iff $s_i > k/2$), and derive the analogous structure.

\begin{quote}
\textbf{Proposition 2.} \emph{Let a $k$-judge panel ($k$ even) decide query $i$ by majority vote with ties broken as ``incorrect.'' Suppose exactly one judge's vote is replaced. Then the panel's decision can change if and only if $s_i \in \{k/2,\ k/2+1\}$.}
\end{quote}

\emph{Proof.} As before, $s_i' \in \{s_i-1, s_i, s_i+1\}$, and decision $=\mathrm{correct}$ iff the tally is $\geq k/2+1$. The boundary lies between the integers $k/2$ (incorrect) and $k/2+1$ (correct), so a $\pm1$ change crosses it only if $s_i \in \{k/2, k/2+1\}$. $\blacksquare$

Unlike the odd case, this set is \emph{not} symmetric in the margin $m_i = |2s_i - k|$: $s_i = k/2$ (a tie) has $m_i=0$, and $s_i = k/2+1$ (the narrowest possible ``correct'' majority) has $m_i=2$ -- but $s_i = k/2-1$, the narrowest possible \emph{incorrect} majority, also has $m_i=2$ and is \emph{not} pivotal under this convention (a further $-1$ vote keeps it below the boundary). Margin alone is thus insufficient to identify pivotal queries once $k$ is even; the \emph{signed} tally relative to $k/2$ is needed, with the two sides of a given margin behaving asymmetrically. We validate this directly in Section~5 by re-analyzing the existing 7-judge data as all $\binom{7}{6}=7$ six-judge (even) subsets, with no new data collection: the theory predicts a structural zero specifically on the incorrect-leaning bare majority ($s_i=2$ for $k=6$), and nowhere else does the odd-panel intuition of ``same margin, same status'' hold.

\paragraph{What requires evidence.} The arithmetic identifies where a single replacement can affect decisions; it does not say how many errors occur there or whether an imperfect signal improves them. We therefore test two empirical questions: whether panel error rate increases as $m_i \to 1$, and whether substitution at $m_i=1$ improves accuracy.

\paragraph{Verification signal and its asymmetry.} For problem $p$ with assertion set $T_p$ and coverage $c$, let $S_c(p) \subseteq T_p$ contain the first $q=\max(1,\min(|T_p|,\mathrm{round}(c|T_p|)))$ assertions in source order. The signal is $V_c(x)=\mathbb{1}[\bigwedge_{t \in S_c(p)}\mathrm{Pass}(x,t)]$. We use $c \in \{5,10,20,50\}\%$ in the main sweep and 13 levels from 2--80\% in Section~5. Because the full-suite pass/fail label uses $T_p$, this signal is a nested proxy, not an independent measurement: a full-suite pass necessarily passes the subset, so the proxy has no false rejections relative to that operational label and errs only by accepting bugs outside $S_c(p)$. This designed one-sided error structure is potentially complementary to the panel's false-rejection skew and limits generalization to other verifiers (Section~7). Replacing source-order selection with uniformly random assertion subsets over 3,115 candidates with usable code and parseable assertions gives $87.10\pm0.12\%$ signal accuracy and $+10.99\pm0.07$pp pivotal gain across five seeds, close to the source-order results (Supplementary~C).

Aggregate correlation gives a mixed diagnosis. On LiveCodeBench, mean signal--judge $\phi$ ($-0.04$) is below mean judge--judge $\phi$ ($0.26$). On HumanEval+/MBPP+, signal--judge $\phi$ ($0.31$) is not lower than judge--judge $\phi$ ($0.28$). The pooled statistic therefore does not establish independence; Section~5 instead measures the different question of margin-conditional decision changes.

A matched-accuracy permutation control further rules out an unqualified independence account (Supplementary~A). The real signal's pivotal gain ($+11.2$pp) is below the mean for 2,000 synthetic signals with identical overall accuracy but randomly permuted error locations ($+14.6$pp; 95\% range $[13.1,16.0]$pp). Both panel accuracy (62.1\% pivotal vs.\ 82.4\% overall) and signal accuracy (81.7\% vs.\ 87.6\%) decline on pivotal queries. Our evidence therefore supports a descriptive mechanism -- high margin-conditional signal accuracy, arising partly from the nested proxy's one-sided errors and lower sensitivity to query difficulty -- but does not isolate statistical independence as its cause.

\paragraph{A worked example.} On HumanEval problem 83 (count $n$-digit integers starting or ending with 1), a \texttt{claude-opus-4-7} solution special-cases $n{=}1$ (returning 1) and otherwise returns the closed form \texttt{18*(10**(n-2))} for $n{>}1$ -- correct by inclusion-exclusion ($10^{n-1}+9\times10^{n-2}-10^{n-2}=18\times10^{n-2}$) -- and it passes the full test suite. Our panel splits 4--3 against it (\texttt{qwen3-max-preview} offers its own mistaken derivation; three judges correctly validate it). Our 20\%-coverage signal, which does not reason about the formula, returns \texttt{CORRECT} and flips the panel: an elaborate, confident, \emph{incorrect} derivation is exactly as persuasive to a plausibility-reasoning judge as a correct one -- a failure mode execution is immune to by construction.

\section{Experimental Setup}

We construct candidates with 7 solver models, execute them in a sandboxed subprocess against the full test suite (the operational pass/fail label) and a limited subset (the signal), and have 3--9 judges -- disjoint from the solver roster -- score each candidate without execution results. Judges state the verdict on the first response line. Judge calls use temperature 0; solver calls use temperature 0 on HumanEval+/MBPP+ and 0.2 on LiveCodeBench. Calls use fixed prompts and a single provider gateway in July 2026. Table~\ref{tab:models} reports exact model IDs; prompts, raw outputs, and data are in the supplementary code/data archive.

The 463 HumanEval+/MBPP+ problems are 163 of HumanEval+'s 164 (HumanEval/32 excluded a priori for a documented evalplus special-oracle bug unrelated to any candidate's correctness) plus a fixed, seeded random sample of 300 of MBPP+'s 378 (oversampled and filtered for unparseable assertions before any candidate was generated) -- both decisions made before any results were observed, not a post-hoc filter.

We test four panel configurations. (1) \textbf{HumanEval+/MBPP+, $k=7$}: 463 problems $\times$ 7 solvers, yielding 3{,}241 candidates with a recorded full-suite outcome and all 7 verdicts. The random-assertion ablation uses 3,115 of them; the remaining 126 lack usable extracted code or a parseable positive assertion count but remain valid full-suite failures for the primary analysis. (2) \textbf{HumanEval+/MBPP+, $k=9$}: the same candidates plus two judges from companies not otherwise in the judge roster. (3) \textbf{$k \in \{3,5\}$}: all 21 five-judge and 35 three-judge subsets of the original panel. These reuse candidates and verdicts and are dependent composition checks, not independent replications. (4) \textbf{LiveCodeBench, $k=7$}: 60 LeetCode-style problems (median 14 tests) $\times$ 7 solvers, with a 70\% full-suite pass rate. HumanEval+/MBPP+ is the discovery set; the other configurations are post-discovery replication checks rather than preregistered confirmatory tests.

For each configuration we compare panel-only accuracy with a policy that samples the replaced judge uniformly. Its expected accuracy is computed exactly by averaging all $k$ leave-one-out substitutions, separately for pivotal ($m=1$) and non-pivotal ($m\geq3$) queries. Section~6 additionally compares fixed and majority-side rules. We report 95\% CIs from 10{,}000 task-level bootstrap resamples.

\begin{table*}[t]
\centering
\small
\setlength{\tabcolsep}{4pt}
\begin{tabular}{@{}rlr@{\qquad}rlr@{}}
\toprule
& Solver ID & Company & & Judge ID & Company \\
\midrule
1 & \texttt{claude-opus-4-7} & Anthropic & 1 & \texttt{gpt-4o-mini} & OpenAI \\
2 & \texttt{gpt-5.4} & OpenAI & 2 & \texttt{claude-haiku-4-5-20251001} & Anthropic \\
3 & \texttt{DeepSeek-V4-Pro} & DeepSeek & 3 & \texttt{DeepSeek-V4-Flash} & DeepSeek \\
4 & \texttt{GLM-5} & Zhipu & 4 & \texttt{qwen3-max-preview} & Alibaba \\
5 & \texttt{Kimi-K2.7-Code} & Moonshot & 5 & \texttt{GLM-5-Turbo} & Zhipu \\
6 & \texttt{doubao-1-5-lite-32k-250115} & ByteDance & 6 & \texttt{Kimi-K2.6} & Moonshot \\
7 & \texttt{gpt-4.1-mini} & OpenAI & 7 & \texttt{Doubao-Seed-1.6} & ByteDance \\
 &  &  & 8 & \texttt{grok-4-1-fast-reasoning} & xAI \\
 &  &  & 9 & \texttt{gemini-2.5-flash-lite} & Google \\
\bottomrule
\end{tabular}
\caption{Model roster. Judges are disjoint from solvers. Judges 8--9 extend the $k{=}7$ panel rather than replacing members and were not selected for accuracy.}
\label{tab:models}
\end{table*}

\section{Results}

\begin{table}[t]
\centering
\small
\setlength{\tabcolsep}{3pt}
\begin{tabular}{@{}lccc@{}}
\toprule
Dataset & Panel & Pivotal gain & Non-pivotal gain \\
\midrule
HE+/MBPP+ & $k{=}7$ & $+11.2$pp [7.7, 14.7] & $0.0000$ \\
HE+/MBPP+ & $k{=}9$\textsuperscript{$\dagger$} & $+10.4$pp [6.3, 14.4] & $0.0000$ \\
LiveCodeBench & $k{=}7$ & $+23.3$pp [18.0, 28.9] & $0.0000$ \\
\bottomrule
\end{tabular}
\caption{Expected accuracy gain under uniformly random single-judge replacement at 20\% test coverage, computed exactly by averaging all replacement choices. Gains are stratified into pivotal ($m{=}1$) and non-pivotal ($m{\geq}3$) queries; brackets are 95\% task-bootstrap CIs. $\dagger$: two newly added, provider-diverse judges, not a resample.}
\label{tab:main}
\end{table}

\begin{figure}[t]
\centering
\includegraphics[width=\linewidth]{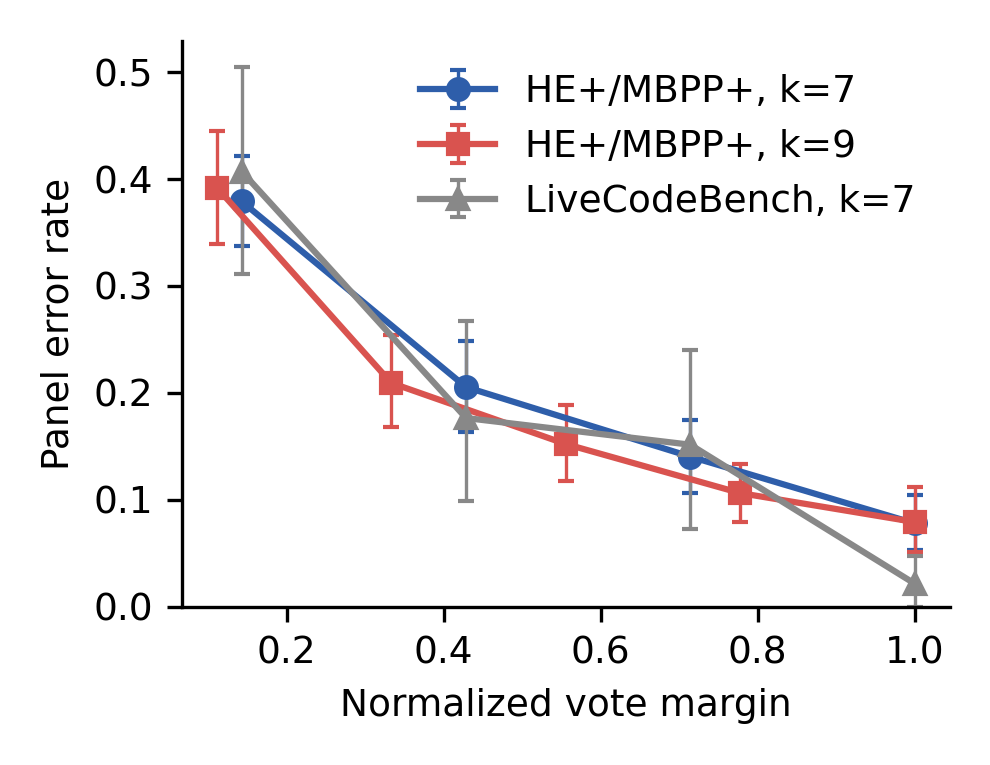}
\caption{Panel error rate by normalized vote margin; bars are 95\% task-bootstrap CIs. Counts from narrowest to widest raw margin are HE+/MBPP+ $k{=}7$: 525/740/1026/950; $k{=}9$: 380/516/769/894/682; LiveCodeBench $k{=}7$: 113/102/66/139.}
\label{fig:margin}
\end{figure}

Table~\ref{tab:main} summarizes the three headline configurations; Figure~\ref{fig:margin} shows that panel error \emph{rate} rises as the vote margin narrows. In every configuration, non-pivotal gain is exactly zero, while pivotal gain is 10--23pp with a bootstrap CI excluding zero ($P(\text{gain}\leq0)<0.001$). This does not mean that most errors are pivotal: in the main setting, 199 errors are pivotal and 370 are non-pivotal. A margin-threshold sweep on $k{=}7$ HumanEval+/MBPP+ gives the same +1.82pp overall gain for every threshold $t\geq1$, because calls outside $m=1$ cannot alter a single-substitution decision.

The $k\in\{3,5\}$ subsampling check (56 dependent judge subsets) reproduces the same pattern: non-pivotal gain is exactly zero and pivotal gain is positive in every subset (Table~\ref{tab:subsample}). On LiveCodeBench, pivotal gain is +23.3pp versus +11.2pp on HumanEval+/MBPP+; this comparison is descriptive because benchmark difficulty and signal accuracy both differ.

\begin{table}[t]
\centering
\small
\begin{tabular}{@{}lccc@{}}
\toprule
Panel size & \#configs & Pivotal gain (range) & Non-pivotal gain \\
\midrule
$k=5$ & 21 & $+7.9$ to $+14.3$pp & $0.0000$ (all 21) \\
$k=3$ & 35 & $+6.5$ to $+16.1$pp & $0.0000$ (all 35) \\
\bottomrule
\end{tabular}
\caption{Dependent composition analysis over every $\binom{7}{5}$ and $\binom{7}{3}$ judge subset from the same $k{=}7$ panel and candidate set. Pivotal gain is positive and non-pivotal gain is zero in all 56 subsets; these are not independent replications.}
\label{tab:subsample}
\end{table}

\paragraph{Aggregate and conditional diagnostics.} Table~\ref{tab:neff} compares the pooled effective-vote count with margin-stratified utility on the same $k{=}7$ HumanEval+/MBPP+ data. The pooled $n_{\mathrm{eff}}$ change is not distinguishable from zero ($-0.04$, 95\% CI $[-0.10,+0.02]$), whereas uniformly random replacement improves overall accuracy by +1.82pp. Stratification locates that average: +11.2pp on the 16.2\% pivotal subset and exactly zero elsewhere. The diagnostics answer different questions -- population-level error dependence versus conditional decision utility.

\begin{table}[t]
\centering
\small
\setlength{\tabcolsep}{3pt}
\begin{tabular}{@{}lccc@{}}
\toprule
Metric & Panel & +signal & Diff [95\% CI] \\
\midrule
$n_{\mathrm{eff}}$ & 2.61 & 2.57 & $-0.04$ [$-0.10$,$+0.02$] \\
Overall acc.\ & 82.44\% & 84.26\% & $+1.82$pp [$+1.22$,$+2.43$] \\
\midrule
Pivotal acc.\ & -- & -- & $+11.2$pp [$+7.7$,$+14.7$] \\
Non-pivotal acc.\ & -- & -- & $0.0000$ \\
\bottomrule
\end{tabular}
\caption{The pooled effective-vote diagnostic ($n_{\mathrm{eff}}$, \citealp{kohli2026nine}) and margin-conditional substitution utility on the same data. Accuracy rows use uniformly random replacement. $k{=}7$, HumanEval+/MBPP+, 20\% test coverage.}
\label{tab:neff}
\end{table}

\paragraph{Cross-benchmark breakdown.} Within $k{=}7$, HumanEval+ ($n{=}236$ pivotal, $905$ non-pivotal) gains $+16.95$pp/$0.0000$; MBPP+ ($n{=}289$/$1811$) gains $+6.52$pp/$0.0000$. The benchmarks differ in magnitude but share the structural zero outside the pivotal region.

\begin{figure}[t]
\centering
\includegraphics[width=\linewidth]{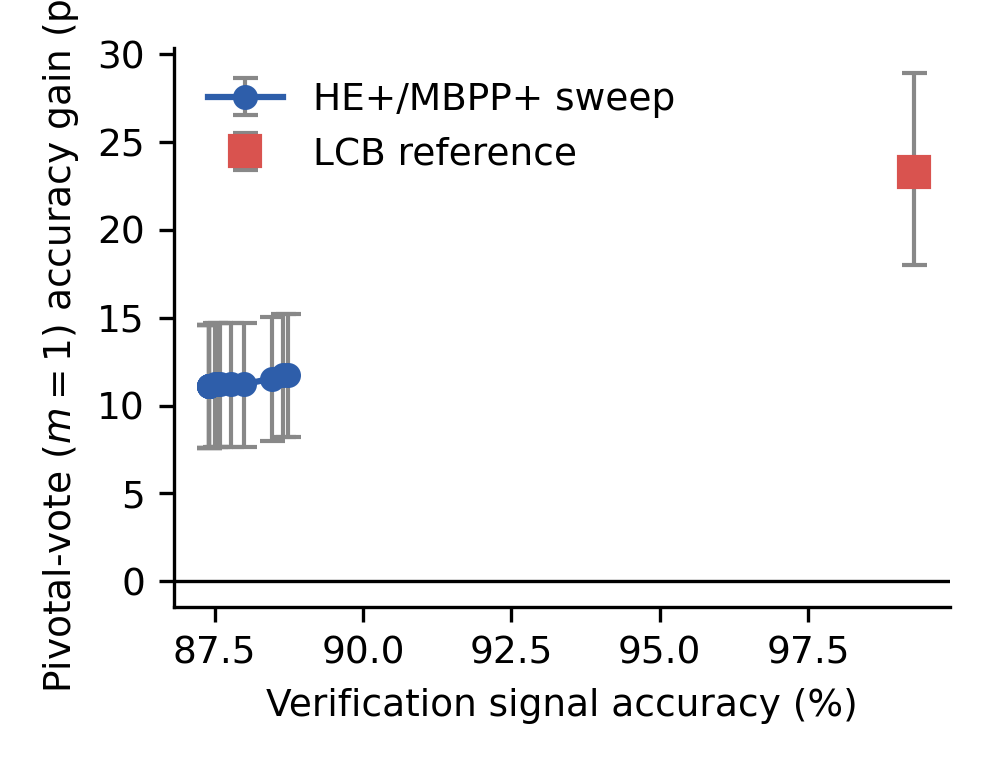}
\caption{Observed pivotal-vote gain over a narrow signal-accuracy range. Blue circles: controlled HumanEval+/MBPP+ sweep with 95\% problem-cluster bootstrap CIs. The unconnected LiveCodeBench point is cross-benchmark context, not evidence of a shared dose-response curve.}
\label{fig:granularity}
\end{figure}

\paragraph{Signal-accuracy sweep.} How does the pivotal-region gain scale with signal accuracy? We re-execute all 3{,}241 candidates locally at 13 test-suite granularities (2--80\% of assertions), yielding signal accuracies 87.4--88.7\%. Figure~\ref{fig:granularity} shows the gain rises monotonically over this narrow range ($+11.10$pp to $+11.76$pp, every CI excluding zero) -- too narrow a span to establish a general dose-response alone. We separately plot the LiveCodeBench point (99.3\% accuracy, $+23.3$pp) for reference only, \emph{not} as evidence the trend extends across benchmarks. A coarse 2\% signal already recovers most of the 80\%-coverage gain, suggesting gating can also be cheap along test-suite depth; lower signal accuracies (below 87\%) are future work (Section~7).

\paragraph{Even panels validate the predicted asymmetry.} We test Proposition~2 by re-deriving all $\binom{7}{6}=7$ six-judge (even) subsets of the original panel, without additional data collection. Table~\ref{tab:even} reports the gain in each region, aggregated across all 7 subsets. The tie region ($s{=}3$) shows the largest gain we test ($+30.1$pp); the bare-\emph{correct}-leaning majority ($s{=}4$, predicted pivotal) shows a smaller, consistently positive gain; critically, the bare-\emph{incorrect}-leaning majority ($s{=}2$, identical $|m_i|{=}2$ to $s{=}4$) shows a gain of \emph{exactly} $0.0000$ in all 7 subsets, exactly as Proposition~2 predicts. This follows directly from the tie-breaking convention specified in the proposition; we are not aware of prior work stating this asymmetry for LLM judge panels, though it is an arithmetic consequence rather than a surprising empirical discovery in itself.

\paragraph{Why panel errors skew toward false rejection near the margin.} Pivotal-region errors are not evenly split: the false-negative rate (correct code voted incorrect) rises from 7.5\% on non-pivotal queries to 33.7\% on pivotal queries ($4.5\times$), while the false-positive rate rises only $1.4\times$ (39.1\% to 53.6\%). Two qualitative hypotheses do \emph{not} explain this (Supplementary~D): no hedging-language difference between error types, and an LLM classification of 100 sampled wrong verdicts finds both equally likely to cite a specific, checkable claim.

\begin{table}[t]
\centering
\small
\setlength{\tabcolsep}{4pt}
\begin{tabular}{@{}lccc@{}}
\toprule
Judge & Reject rate & Accuracy & MCC \\
\midrule
GLM-5-Turbo & 10.5\% & 87.2\% & 0.541 \\
Kimi-K2.6 & 23.5\% & 80.4\% & 0.426 \\
Claude Haiku & 20.9\% & 79.0\% & 0.349 \\
Qwen3-max & 29.2\% & 72.8\% & 0.280 \\
DeepSeek-V4-Flash & 33.1\% & 73.7\% & 0.359 \\
GPT-4o-mini & 42.5\% & 64.2\% & 0.247 \\
Doubao-Seed-1.6 & 52.6\% & 58.5\% & 0.253 \\
\bottomrule
\end{tabular}
\caption{Judges' reject rate, accuracy, and Matthews correlation coefficient (MCC), a base-rate-robust discrimination metric. The full-suite operational label is 80.4\% correct.}
\label{tab:calibration}
\end{table}

Part of this is compositional rather than purely psychological. Table~\ref{tab:calibration} helps separate a mechanical effect from a genuine one. Raw accuracy's correlation with reject rate ($r{=}{-}0.99$) is inflated by the 80.4\% base rate: leniency is rewarded almost by construction. Base-rate-invariant balanced accuracy correlates with reject rate far more weakly ($r{=}{-}0.61$), confirming part of the gap is this artifact. But MCC, also base-rate-robust, still correlates with reject rate at $r{=}{-}0.85$ and with panel accuracy across all 28 real subsets at $r{=}0.86/0.85$ (vs.\ $r{=}0.94/0.97$ for raw accuracy; Supplementary~D) -- a genuine, smaller discrimination gap survives alongside the mechanical one. This panel includes more strict, lower-MCC judges than lenient ones, so its aggregate error where votes are close enough for individual idiosyncrasy to decide skews toward their shared error, false rejection -- consistent with a compositional explanation for the correlated-error problem motivating this paper (\citealp{kohli2026nine}), paralleling annotator calibration in crowdsourcing (\citealp{dawid1979maximum}), though with only 7 judges we do not claim an isolated causal mechanism. A cheap pre-deployment check -- confusion matrix and MCC on a calibration sample -- can flag both effects before they distort the votes that matter most.

\begin{table}[t]
\centering
\small
\begin{tabular}{@{}lccc@{}}
\toprule
Region ($k{=}6$) & Status & Gain & Range \\
\midrule
Tie ($s{=}3$) & pivotal & $+30.1$pp & $[26.9, 32.6]$ \\
$s{=}4$ & pivotal & $+0.8$pp & $[0.3, 1.5]$ \\
$s{=}2$ & non-pivotal & $0.0000$ & all 7 exact \\
$s{\leq}1$ or $s{\geq}5$ & non-pivotal & $0.0000$ & all 7 exact \\
\bottomrule
\end{tabular}
\caption{Accuracy gain across all $\binom{7}{6}=7$ six-judge subsets, by region. Despite $s{=}2$ and $s{=}4$ sharing the same $|m_i|=2$, only $s{=}4$ is pivotal -- confirming Proposition~2's asymmetry prediction.}
\label{tab:even}
\end{table}

\section{Discussion}

\paragraph{Aggregate independence statistics do not characterize conditional utility.} The effective-votes metric of \citet{kohli2026nine} diagnoses correlated errors, averaged over the full query population; it was not designed to detect margin-conditional decision utility, and Table~\ref{tab:neff} shows directly that it does not. When a signal's value is concentrated in a minority subset (12--27\% here) and exactly zero elsewhere, the average can look statistically insignificant even though the underlying effect is large -- without the statistic being wrong for the question it was built to answer. We suspect this generalizes: any intervention whose value is conditional on an observable query property can look ineffective under a non-stratifying aggregate statistic. We recommend reporting effects conditional on panel disagreement alongside any aggregate independence statistic.

\paragraph{Margin gating is separate from replacement choice.} Algorithm~\ref{alg:gated} accepts a specified replacement rule $R$. Proposition~1 guarantees that applying $R$ only at $m=1$ produces the same decisions as applying the same single-ballot rule universally; it does not say that all choices of $R$ are equally accurate. Tables~\ref{tab:main}--\ref{tab:even} report the exact expectation for uniform-random replacement. Fixed-index accuracy ranges from 82.6--85.4\%, while a majority-side rule reaches 85.62\% (Table~\ref{tab:baselines}). Thus gating identifies \emph{when} to call the signal; the replacement rule determines how strongly that signal influences a pivotal vote.

\begin{algorithm}[t]
\caption{Margin-Gated Verification (odd $k$)}
\label{alg:gated}
\begin{algorithmic}[1]
\STATE \textbf{Input:} query $x$, odd-sized $k$-judge panel $J = (j_1,\dots,j_k)$, signal $V$, specified single-ballot replacement rule $R$
\STATE Collect votes $v_1,\dots,v_k \gets J(x)$; let $s \gets \sum_l v_l$
\STATE $m \gets |2s - k|$
\IF{$m = 1$}
    \STATE $v^* \gets V(x)$; $r \gets R(v_1,\dots,v_k)$
    \STATE $v_r \gets v^*$ \COMMENT{$R$: fixed, uniform-random, or majority-side}
    \STATE \textbf{return} $\mathbb{1}[\sum_l v_l > k/2]$
\ELSE
    \STATE \textbf{return} $\mathbb{1}[s > k/2]$ \COMMENT{$V(x)$ provably cannot change this output (Prop.~1)}
\ENDIF
\end{algorithmic}
\end{algorithm}

\paragraph{Alternative deployment policies.} Table~\ref{tab:baselines} separates query gating from replacement choice. Uniform-random replacement improves panel accuracy from 82.44\% to 84.26\%; majority-side replacement reaches 85.62\% (+3.18pp, 95\% CI [2.10, 4.26]) at the same 16.2\% call rate. On a pivotal vote, this rule lets the signal overturn the current majority when they disagree; it is decision-equivalent to giving the signal two votes universally, but avoids non-pivotal calls. Signal-only (87.60\%) and the best judge (87.16\%) remain stronger, so the deployment claim is conditional on retaining a panel. Retention can be an external constraint rather than an accuracy choice -- for example, an audit or governance protocol may require multi-provider verdicts, or a deployment may need tolerance to one judge's outage or model drift. We do not test those benefits; our policy claim applies only after such a constraint has been imposed.

\begin{table}[t]
\centering
\small
\setlength{\tabcolsep}{3pt}
\begin{tabular}{@{}lcc@{}}
\toprule
Policy & Verifier calls & Accuracy \\
\midrule
Panel only & 0\% & 82.44\% \\
Best judge & 0\% & 87.16\% \\
Signal only & 100\% & 87.60\% \\
Universal, uniform repl. & 100\% & 84.26\% \\
Gate, uniform repl. & 16.2\% & 84.26\% \\
Gate, majority-side & 16.2\% & 85.62\% \\
Random-query gate & 16.2\% & 82.74\%$\pm$0.09 \\
\bottomrule
\end{tabular}
\caption{Policy comparison, $k{=}7$ HumanEval+/MBPP+. Gating preserves a specified substitution rule at $6\times$ fewer calls. Replacement choice matters: majority-side exceeds uniform replacement, while signal-only and the best judge remain stronger.}
\label{tab:baselines}
\end{table}

Sandboxed execution takes 49ms/call vs.\ 5.72s for a judge call (Supplementary~C). Universal verification is therefore already cheap in our setup; gating matters economically when the verifier is expensive, not for this specific signal.

\section{Limitations}

The signal is a nested subset of the full-suite operational label, giving it one-sided errors: it can miss uncovered bugs but cannot reject a full-suite pass. Its complementarity with the panel's false-rejection skew may therefore not transfer to independent, symmetric, or abstaining verifiers. Signal accuracy is high (87--99\%), and the controlled sweep spans only 87.4--88.7\%; behavior below 87\% remains unknown. Signal-only and the best judge outperform every panel policy tested, so gating is relevant only when a panel must be retained. Replacement-policy comparisons are post-hoc on the discovery data and need held-out validation. LiveCodeBench has 113 pivotal observations; all benchmarks are code tasks. API aliases may drift, and partly overlapping providers introduce a measured same-provider effect. Weighted panels, non-binary signals, and non-code domains remain open.

\section{Reproducibility and Ethics}

No human subjects or personal data are involved. Untrusted candidate code runs under a danger-pattern scan, timeout, and process isolation. Section~4 reports model IDs, sample sizes, and bootstrap procedures. CIs capture task-sampling rather than call-level uncertainty: 3 of 15 sampled pivotal query--judge groups produced at least one different verdict across five repeated calls (Supplementary~C).

\section{Conclusion}

For single-ballot substitution in an unweighted-majority panel, only pivotal decisions can change. Across three code benchmarks and four panel sizes, substitution gain is concentrated there and zero elsewhere. Aggregate dependence and margin-conditional utility therefore answer different questions. When retaining a panel, gate a specified replacement rule on pivotal votes; in our main setting, majority-side replacement is stronger than uniform replacement, although simpler signal-only and single-judge policies remain stronger still.

\bibliographystyle{plainnat}
\bibliography{references}

@misc{kohli2026nine,
  title={Nine Judges, Two Effective Votes: Correlated Errors Undermine LLM Evaluation Panels},
  author={Kohli, Guneet},
  year={2026},
  eprint={2605.29800},
  archivePrefix={arXiv},
  primaryClass={cs.CL}
}

@misc{kuai2026independent,
  title={How Independent are Large Language Models? A Statistical Framework for Auditing Behavioral Entanglement and Reweighting Verifier Ensembles},
  author={Kuai, Chenchen and Jiang, Jiwan and Zhu, Zihao and Wang, Hao and Wu, Keshu and Li, Zihao and Zhang, Yunlong and Liu, Chenxi and Tu, Zhengzhong and Fan, Zhiwen and Zhou, Yang},
  year={2026},
  eprint={2604.07650},
  archivePrefix={arXiv},
  primaryClass={cs.CL}
}

@misc{kwok2026llmverifier,
  title={LLM-as-a-Verifier: A General-Purpose Verification Framework},
  author={Kwok, Jacky and Li, Shulu and Atreya, Pranav and Liu, Yuejiang and Jiang, Yixing and Finn, Chelsea and Pavone, Marco and Stoica, Ion and Mirhoseini, Azalia},
  year={2026},
  eprint={2607.05391},
  archivePrefix={arXiv},
  primaryClass={cs.CL}
}

@misc{akinfaderin2026fregelogic,
  title={FregeLogic at SemEval 2026 Task 11: A Hybrid Neuro-Symbolic Architecture for Content-Robust Syllogistic Validity Prediction},
  author={Akinfaderin, Adewale and Diallo, Nafi},
  year={2026},
  eprint={2604.18328},
  archivePrefix={arXiv},
  primaryClass={cs.CL}
}

@article{banzhaf1965weighted,
  title={Weighted Voting Doesn't Work: A Mathematical Analysis},
  author={Banzhaf, John F.},
  journal={Rutgers Law Review},
  volume={19},
  pages={317--343},
  year={1965}
}

@misc{condorcet1785essai,
  title={Essai sur l'Application de l'Analyse \`a la Probabilit\'e des D\'ecisions Rendues \`a la Pluralit\'e des Voix},
  author={Condorcet, Marquis de},
  year={1785},
  howpublished={Paris: Imprimerie Royale}
}

@misc{chen2021evaluating,
  title={Evaluating Large Language Models Trained on Code},
  author={Chen, Mark and Tworek, Jerry and Jun, Heewoo and Yuan, Qiming and Pinto, Henrique Ponde de Oliveira and Kaplan, Jared and Edwards, Harri and Burda, Yuri and Joseph, Nicholas and Brockman, Greg and Ray, Alex and Puri, Raul and Krueger, Gretchen and Petrov, Michael and Khlaaf, Heidy and Sastry, Girish and Mishkin, Pamela and Chan, Brooke and Gray, Scott and Ryder, Nick and Pavlov, Mikhail and Power, Alethea and Kaiser, Lukasz and Bavarian, Mohammad and Winter, Clemens and Tillet, Philippe and Such, Felipe Petroski and Cummings, Dave and Plappert, Matthias and Chantzis, Fotios and Barnes, Elizabeth and Herbert-Voss, Ariel and Guss, William Hebgen and Nichol, Alex and Paino, Alex and Tezak, Nikolas and Tang, Jie and Babuschkin, Igor and Balaji, Suchir and Jain, Shantanu and Saunders, William and Hesse, Christopher and Carr, Andrew N. and Leike, Jan and Achiam, Josh and Misra, Vedant and Morikawa, Evan and Radford, Alec and Knight, Matthew and Brundage, Miles and Murati, Mira and Mayer, Katie and Welinder, Peter and McGrew, Bob and Amodei, Dario and McCandlish, Sam and Sutskever, Ilya and Zaremba, Wojciech},
  year={2021},
  eprint={2107.03374},
  archivePrefix={arXiv},
  primaryClass={cs.LG}
}

@misc{austin2021program,
  title={Program Synthesis with Large Language Models},
  author={Austin, Jacob and Odena, Augustus and Nye, Maxwell and Bosma, Maarten and Michalewski, Henryk and Dohan, David and Jiang, Ellen and Cai, Carrie and Terry, Michael and Le, Quoc and Sutton, Charles},
  year={2021},
  eprint={2108.07732},
  archivePrefix={arXiv},
  primaryClass={cs.PL}
}

@inproceedings{liu2023your,
  title={Is Your Code Generated by ChatGPT Really Correct? Rigorous Evaluation of Large Language Models for Code Generation},
  author={Liu, Jiawei and Xia, Chunqiu Steven and Wang, Yuyao and Zhang, Lingming},
  booktitle={Advances in Neural Information Processing Systems 36 (NeurIPS)},
  year={2023}
}

@inproceedings{jain2024livecodebench,
  title={LiveCodeBench: Holistic and Contamination Free Evaluation of Large Language Models for Code},
  author={Jain, Naman and Han, King and Gu, Alex and Li, Wen-Ding and Yan, Fanjia and Zhang, Tianjun and Wang, Sida and Solar-Lezama, Armando and Sen, Koushik and Stoica, Ion},
  booktitle={International Conference on Learning Representations (ICLR)},
  year={2025}
}

@inproceedings{seung1992query,
  title={Query by committee},
  author={Seung, H. S. and Opper, Manfred and Sompolinsky, Haim},
  booktitle={Proceedings of the Fifth Annual Workshop on Computational Learning Theory (COLT)},
  pages={287--294},
  year={1992}
}

@inproceedings{geifman2017selective,
  title={Selective Classification for Deep Neural Networks},
  author={Geifman, Yonatan and El-Yaniv, Ran},
  booktitle={Advances in Neural Information Processing Systems (NeurIPS)},
  volume={30},
  year={2017},
  eprint={1705.08500},
  archivePrefix={arXiv}
}

@inproceedings{mozannar2020consistent,
  title={Consistent Estimators for Learning to Defer to an Expert},
  author={Mozannar, Hussein and Sontag, David},
  booktitle={Proceedings of the 37th International Conference on Machine Learning (ICML)},
  year={2020},
  eprint={2006.01862},
  archivePrefix={arXiv}
}

@misc{verga2024replacing,
  title={Replacing Judges with Juries: Evaluating LLM Generations with a Panel of Diverse Models},
  author={Verga, Pat and Hofstatter, Sebastian and Althammer, Sophia and Su, Yixuan and Piktus, Aleksandra and Arkhangorodsky, Arkady and Xu, Minjie and White, Naomi and Lewis, Patrick},
  year={2024},
  eprint={2404.18796},
  archivePrefix={arXiv},
  primaryClass={cs.CL}
}

@article{dawid1979maximum,
  title={Maximum Likelihood Estimation of Observer Error-Rates Using the {EM} Algorithm},
  author={Dawid, A. P. and Skene, A. M.},
  journal={Journal of the Royal Statistical Society: Series C (Applied Statistics)},
  volume={28},
  number={1},
  pages={20--28},
  year={1979}
}

@article{chen2023frugalgpt,
  title={FrugalGPT: How to Use Large Language Models While Reducing Cost and Improving Performance},
  author={Chen, Lingjiao and Zaharia, Matei and Zou, James},
  journal={Transactions on Machine Learning Research},
  year={2024}
}

\end{document}